\documentclass[preprint]{elsarticle}
\usepackage{graphicx,amssymb,mathrsfs,amsmath,pifont,amscd,latexsym,amsfonts,amsthm,colortbl,bm}
\usepackage{algpseudocode}
\usepackage[linesnumbered,ruled]{algorithm2e}
\usepackage{geometry}
\usepackage{graphics}
\usepackage{epstopdf}
\usepackage{float}
\usepackage{subfigure}
\usepackage{epsfig}
\usepackage{setspace}
\usepackage{booktabs}
\usepackage{xcolor}
\usepackage{multirow}
\usepackage{lineno}
\usepackage{color}
\usepackage[colorlinks,urlcolor=blue,driverfallback=dvipdfm]{hyperref}

\graphicspath{{figures/}}

\begin{document}
\begin{frontmatter}

\title{Joint Super-Resolution and Segmentation for 1-m Impervious Surface Area Mapping in China’s Yangtze River Economic Belt}

\author[secondaddress]{Jie Deng}
\ead{dengjie@aircas.ac.cn}

\author[secondaddress,thirdaddress]{Danfeng Hong}
\ead{hongdf@aircas.ac.cn}

\author[fourthaddress]{Chenyu Li}
\ead{chenyuli.erh@gmail.com}

\author[fifthaddress,sixthaddress]{Naoto Yokoya}
\ead{yokoya@k.u-tokyo.ac.jp}

\address[secondaddress]{Aerospace Information Research Institute, Chinese Academy of Sciences, Beijing 100094, China;}
\address[thirdaddress]{School of Electronic, Electrical and Communication Engineering, University of Chinese Academy of Sciences, Beijing 100049, China;}
\address[fourthaddress]{School of Mathematics, Southeast University, Nanjing 210096, China;}
\address[fifthaddress]{Graduate School of Frontier Sciences, the University of Tokyo, Chiba 277-8561, Japan;}
\address[sixthaddress]{RIKEN Center for Advanced Intelligence Project, Tokyo 103-0027, Japan.}

\begin{abstract}
High-resolution impervious surface area (ISA) mapping is critical for a wide range of applications, including sustainable urban planning, flood risk assessment, and land use monitoring. However, the production of meter ISA products has long relied on commercial very high-resolution (VHR) satellite imagery, which is cost-prohibitive and geographically limited. To this end, we propose a novel joint framework by integrating super-resolution and segmentation, called JointSeg, which enables the generation of 1-meter ISA maps directly from freely available Sentinel-2 imagery (10m resolution). JointSeg was trained on multimodal cross-resolution inputs, offering a scalable and affordable alternative to traditional approaches. This synergistic design enables gradual resolution enhancement from 10m to 1m while preserving fine-grained spatial textures, and ensures high classification fidelity through effective cross-scale feature fusion. This method has been successfully applied to the Yangtze River Economic Belt (YREB), a region characterized by complex urban-rural patterns and diverse topography. As a result, a comprehensive ISA mapping product for 2021, referred to as ISA-1, was generated, covering an area of over 2.2 million square kilometers. Quantitative comparisons against the 10m ESA WorldCover and other benchmark products reveal that ISA-1 achieves an F1-score of 85.71\%, outperforming bilinear-interpolation-based segmentation by 9.5\%, and surpassing other ISA datasets by 21.43\%-61.07\%. In densely urbanized areas (e.g., Suzhou, Nanjing), ISA-1 reduces ISA overestimation through improved discrimination of green spaces and water bodies. Conversely, in mountainous regions (e.g., Ganzi, Zhaotong), it identifies significantly more ISA due to its enhanced ability to detect fragmented anthropogenic features such as rural roads and sparse settlements, demonstrating its robustness across diverse landscapes. Moreover, we present biennial ISA maps from 2017 to 2023, capturing spatiotemporal urbanization dynamics across representative cities. The results highlight distinct regional growth patterns: rapid expansion in upstream cities, moderate growth in midstream regions, and saturation in downstream metropolitan areas. These findings underscore the framework’s utility for continuous, high-resolution urban monitoring using only open-access data. In summary, our method establishes a transformative approach to ISA mapping by overcoming the resolution limitations of medium-resolution imagery. This technique opens new possibilities for large-scale, fine-resolution ISA monitoring without the dependency on expensive VHR data, thus supporting global efforts in sustainable development and urban resilience planning.
\end{abstract}


\end{frontmatter}
\section{Introduction}

The YREB is a key national development strategy in China, covering 11 provinces and municipalities across the eastern, central, and western regions. It spans 27.7\% of the country’s land area and is home to over 44.2\% of the population, contributing 47.7\% of the GDP \cite{China_2024YREB,jin2018spatiotemporal}. This region is characterized by diverse terrains and ecosystems, from the Qinghai-Tibet Plateau to the Yangtze River Delta. It is a key area for the coordinated development of the economy and ecology. Rapid urbanization and industrialization have led to a significant increase in ISA, including roads, buildings, squares, and other artificial impervious areas. These changes have profoundly impacted regional ecological and biogeochemical cycles, hydrological processes, urban heat island effects, biodiversity, and carbon fluxes in terrestrial ecosystems \cite{li2022impacts}. ISA is a direct indicator of urbanization, providing spatial data that reflects human activities. Accurately mapping the spatiotemporal dynamics of ISA in the YREB can reveal urbanization patterns and offer insights into its multifaceted impacts on climate change, environmental transformation, and sustainable development. This is crucial for monitoring urban expansion, population shifts, economic growth, and terrestrial ecosystem services \cite{huang2022mapping}. 

Mid-resolution satellite imagery (e.g., Landsat, Sentinel) has enabled large-scale temporal-spatial mapping of ISA across regional to global scales, resulting in 30-m resolution products such as GAIA \cite{gong2020annual}, GISA\cite{WOS:000679271200001}, GISD30\cite{zhang2021gisd30}, NUACI\cite{WOS:000430897300017} and GAUD\cite{WOS:000530241600002}, alongside 10-m products like GHSL-2018\cite{WOS:000584647300001}, GISA-10m\cite{huang2022mapping}, Hi-GISA\cite{WOS:000798908800001}, ESA WorldCover \cite{zanaga2022esa} and ESRI Land Use/Land Cover\cite{karra2021global}. While mid-resolution data balance spectral and spatial capabilities, their limited pixel resolution hampers accurate detection of ISA's fine-grained spatial patterns. For instance, urban tree crowns generally range from 0.5 to 10 meters in diameter, often falling below the spatial resolution of Sentinel-2 imagery. Consequently, these features may be spectrally mixed with adjacent impervious surfaces, thereby contributing to a potential overestimation of ISA\cite{WOS:000741335600001}. The YREB,  with its complex topography spanning multiple geomorphological units - the Qinghai-Tibet Plateau, Yunnan-Guizhou Plateau, Sichuan Basin, and Middle-Lower Yangtze Plain - features a mosaic of mountainous, hilly, and flat terrains. Moreover, its highly diversified land-use structures, with intertwined spatial patterns, elevate the need for finer-resolution ISA mapping. Coarse resolution products risk distorting urban morphology \cite{WOS:000367105900004,li2022impacts,WOS:000274058100006}, further compounded by the findings of Chakraborty et al.\cite{WOS:001342098500015}, which highlight significant discrepancies among global urban datasets across spatiotemporal scales, emphasizing the limitations of coarse-resolution products in accurately capturing urban dynamics.





The 1 m resolution ISA dataset serves as an established benchmark for spatial statistical analysis\cite{li2022impacts}, outperforming coarse-resolution imagery in resolving urban microstructures, peri-urban transition dynamics, and fine-scale landscape heterogeneity\cite{WOS:000997229500001}. The high-fidelity products elucidate complex spatial patterns and functional urban characteristics\cite{WOS:000812326900002, WOS:000549189200044}. Traditional high-resolution mapping relies on costly VHR images (e.g., WorldView, Gaofen, and UAV imagery), which provide rich textural information that allows for precise differentiation between built structures and vegetation\cite{WOS:000975480700002, WOS:001296699100004, WOS:000812326900002,WOS:000363666900006}. While open-access VHR archives (e.g., Google, Microsoft) help reduce costs, their irregular revisit cycles and sparse historical coverage constrain ISA monitoring. Sentinel-2 and similar medium-resolution images offer a practical alternative, providing enhanced spatiotemporal accessibility. Advances in AI-based super-resolution (SR) techniques applied to these open datasets have the potential to overcome the dual limitations of VHR imagery (restricted coverage and infrequent updates) while addressing the constraints of native medium-resolution imagery, thus democratizing high-precision ISA mapping.

Recent years have witnessed significant advancements in SR techniques, particularly driven by breakthroughs in deep learning and AI \cite{WOS:001269395000002,WOS:000739651101110,sirko2023high}. Numerous studies have integrated SR approaches to enhance the spatial resolution of low-resolution imagery, enabling large-scale acquisition of high-detail images and mapping products. For instance, Feng et al. \cite{WOS:000964238600001} mapped 2.5 m-resolution building footprints across 35 Chinese cities using Sentinel-2 data. Cao and Weng \cite{WOS:001257342900001} developed a deep learning-based SR framework to generate 2.5 m-resolution building height maps from Sentinel-1/2 imagery. However, current SR implementations in remote sensing predominantly achieve a 4× resolution enhancement (e.g., upscaling from 10m to 2.5m), while meter-scale targets (e.g., 10× upsampling from 10m to 1m) face critical challenges, such as excessive smoothing and texture distortion due to the loss of high-frequency detail. In contrast, Generative Adversarial Network (GAN)-based SR algorithms (e.g., Real-ESRGAN) exhibit superior performance in handling severely degraded real-world imagery. Validations using Landsat and Sentinel-2 data confirm GANs' effectiveness \cite{gupta2024senglean,WOS:000783587500001,WOS:000999695000001}. However, their potential for generating meter-scale ISA mapping products with stringent geometric and semantic accuracy requirements remains underexplored. 

Moreover, accurate annual ISA (Impervious Surface Area) mapping products offer an intuitive means to understand the spatiotemporal dynamics of urban development, supporting municipal authorities in policy-making, ecological monitoring, and land-use planning. However, traditional high-resolution ISA mapping efforts that rely on meter-level remote sensing imagery have rarely explored long-term time series \cite{WOS:001171605700001,WOS:000812326900002} primarily due to the difficulty in acquiring very high-resolution data at regular intervals. In contrast, the data source used for our meter-scale ISA products—Sentinel-2 imagery—is globally and consistently available, offering substantial potential for producing near-real-time, annual ISA maps at meter-level spatial resolution.

In summary, this study presents a novel technical framework for large-scale, high-accuracy, and long-term ISA mapping, based on a deep learning-driven super-resolution segmentation approach using openly accessible Sentinel-2 imagery. The major innovations of this research are threefold:
\begin{itemize}
    \item A large-scale ISA super-resolution segmentation dataset (ISASeg) was constructed, consisting of 10-meter Sentinel-2 images, 1-meter super-resolved imagery, and corresponding 1-meter ISA labels. The dataset contains over 4.26 billion labeled ISA pixels, significantly advancing the research and application of fine-scale ISA mapping over large areas.
    \item A progressive generative adversarial network (ProgESRGAN) was proposed for the super-resolution reconstruction of Sentinel-2 imagery, achieving a 10 m-to-1 m enhancement with rich structural details. This output was subsequently integrated with Mask2Former to perform super-resolution semantic segmentation, enabling ISA mapping at 1m resolution directly from 10m Sentinel-2 imagery, effectively breaking through the resolution limitations of traditional mapping methods.
    \item High-resolution (1m) ISA maps were generated across the entire Yangtze River Economic Belt, and long-term ISA mapping products (2017–2023, biennial) were produced for six representative cities across the upstream, midstream, and downstream regions. These products reveal the spatiotemporal evolution of impervious surfaces and provide valuable high-resolution data to support urban expansion monitoring, ecological assessment, and territorial spatial planning.
\end{itemize}




\section{Dataset and Methods}

\begin{figure*}[!t]
\centering
\includegraphics[width=1\textwidth]{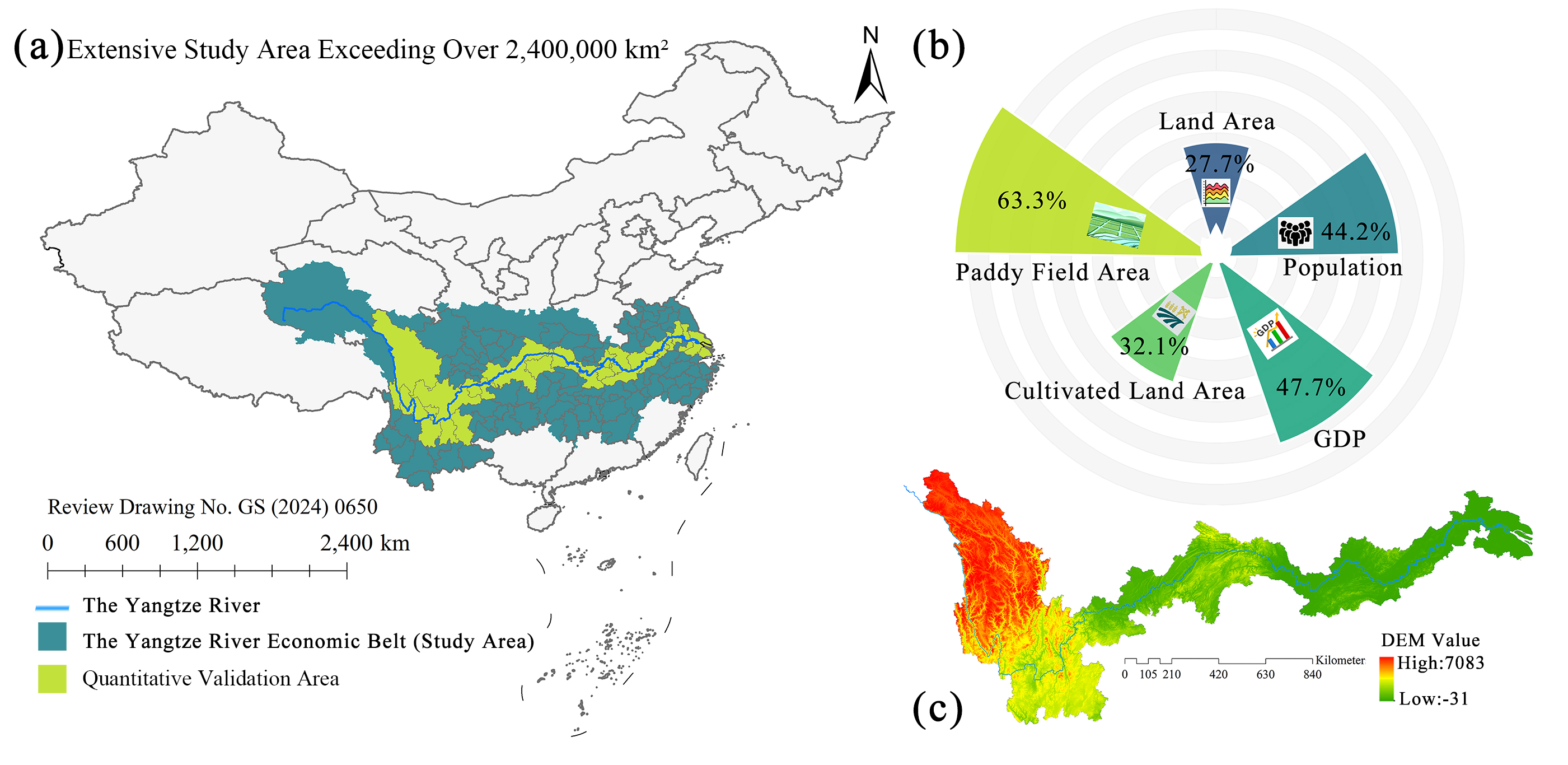}
\caption{\centering (a)The Yangtze River Economic Belt (YREB, shown in blue) is one of the most urbanized and economically dynamic regions in the world. Between 2010 and 2020, its urban population grew by 109 million, and its GDP increased by 163.8\% \cite{China_2024YREB}. (b)The YREB covers 27.7\% of China’s land, hosts 44.2\% of its population, contributes 47.7\% of GDP, and contains 32.1\% of arable land and 63.3\% of paddy fields \cite{China_2024YREB}. (c)The elevation area covers an elevation gradient from 31 meters to 7,083 meters and a total area of over 720,000 square kilometers. }
\label{fig:fig1}
\end{figure*}

\subsection{Study area}
The YREB is one of the world's most densely populated and industrialized economic regions, experiencing rapid urbanization compared with other parts of China. Between 2010 to 2020, the urban population in the YREB grew by 109 million, while its total economic output surged by 163.8\%\cite{China_2024YREB}. Covering 27.7\% of China's land area, the region is home to 44.2\% of the national population, contributes 47.7\% of mainland China's GDP, and contains 32.1\% of the country’s arable land and 63.3\% of its paddy fields\cite{China_2024YREB}. 

To evaluate the reliability of the proposed method in generating fine-scale ISA products across large-scale regions, we produced ISA maps for the entire Yangtze River Economic Belt (Fig.\ref{fig:fig1}, red region), which encompasses all administrative areas of 11 provincial-level units in China, as well as relevant county-level divisions from an additional 8 provinces. In total, the mapping covers 1,173 county-level administrative units, with a combined land area exceeding 2.4 million square kilometers.

To further assess the accuracy and consistency of the generated products, we conducted a detailed quantitative analysis across 36 prefecture-level cities situated along the main stem of the Yangtze River (Fig. \ref{fig:fig1}, orange region). These cities range from megacities such as Shanghai, Wuhan, Nanjing, and Chongqing to less-developed regions such as Chuxiong Yi Autonomous Prefecture, Diqin Tibetan Autonomous Prefecture, and Enshi Tujia and Miao Autonomous Prefecture, spanning an elevation gradient from 31 to 7,083 meters, and covering a total area of over 720,000 square kilometers.

In addition, to demonstrate the capability of our approach for multi-year, near-real-time ISA mapping, we generated biennial ISA products (2017, 2019, 2021 and 2023) for six representative cities located in the upstream (Chongqing, Chengdu), midstream (Wuhan, Changsha), and downstream (Nanjing, Shanghai) sections of the Yangtze River Basin. A series of statistical analyses were performed to investigate the urbanization trajectories across different subregions of the Yangtze River Economic Belt.

\begin{figure*}[!t]
\centering
\includegraphics[width=1\textwidth]{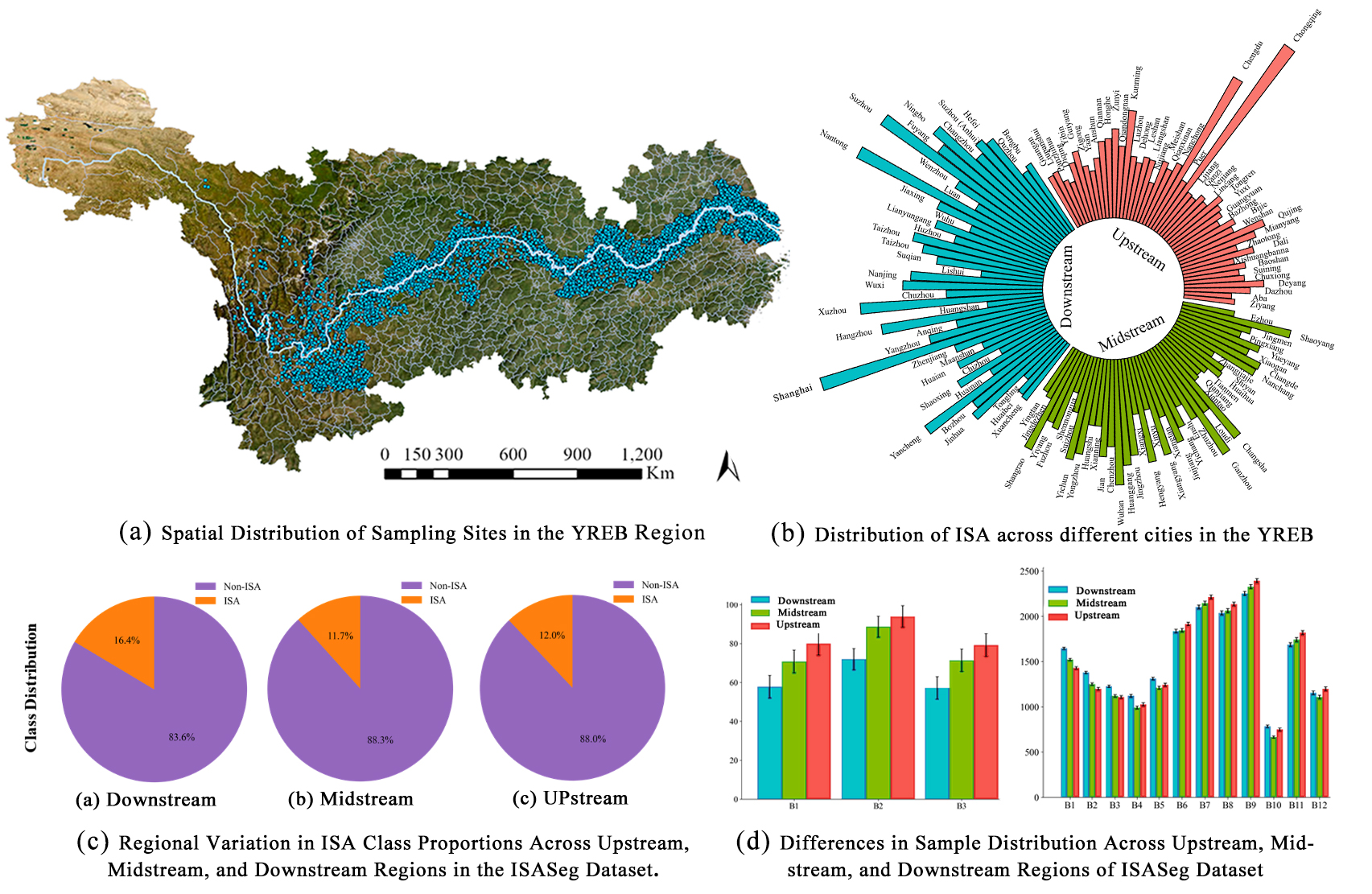}
\caption{\centering Distribute sample points within the study area while ensuring that the minimum distance between any two points is 3 km to prevent overlap, and provide statistics for the ISASeg dataset.}
\label{fig:fig2}
\end{figure*}

\subsection{ISA Super-resolution segmentation dataset (ISASeg)}
This study focused on ISA detection in the YREB, a region characterized by mountainous terrain in its upper reaches and densely distributed paddy fields in the middle-lower reaches. To address spatial heterogeneity, we developed a multi-scale sampling strategy based on built-up area classifications from the ESA land cover product. A stratified random sampling approach was implemented across three subregions (upstream, midstream, and downstream), with sampling points spaced at minimum intervals of 3 km to ensure spatial independence. Subsequently, high-resolution (HR) and low-resolution (LR) satellite imagery were systematically acquired alongside corresponding ISA labels, forming a multi-source remote sensing training dataset specifically tailored to the region's surface characteristics (Fig. \ref{fig:fig2}). 

\noindent\textbf{Low-resolution RGB data (LR):} Sentinel-2 provides 10 m-resolution multispectral imagery with global coverage and a five-day revisit cycle at the equator. Its balanced spatial, spectral, and temporal resolutions make it well-suited for dynamic ISA mapping. Using Google Earth Engine (GEE), we generated 2021 annual composites (median values)  for the study area, excluding scenes with >20\% cloud coverage to minimize atmospheric interference in the YREB. Only the RGB bands were retained to match the spectral channels of high-resolution reference imagery.The imagery retains all 12 original spectral bands from Sentinel-2, including B1, B2, B3, B4, B5, B6, B7, B8, B8A, B9, B11, and B12.

\noindent\textbf{High-resolution RGB data (HR):} We collected the 0.75m-resolution RGB tiles in 2022 from Jilin-1 satellite constellation (\href{www.jl1mall.com}{www.jl1mall.com}) as HR target images. Using ESA WorldCover maps to guide systematic sampling of ISA (built-up areas), we extracted 11,559 non-overlapping 1600$\times$1600 pixel patches (resampled to 1m resolution) from the Jilin-1 imagery suite via QGIS. These HR images were then co-registered with spatially corresponding Sentinel-2 LR patches (10m resolution, 160$\times$160 pixels) to create matched training pairs.

\begin{figure*}[!t]
\centering
\includegraphics[width=1\textwidth]{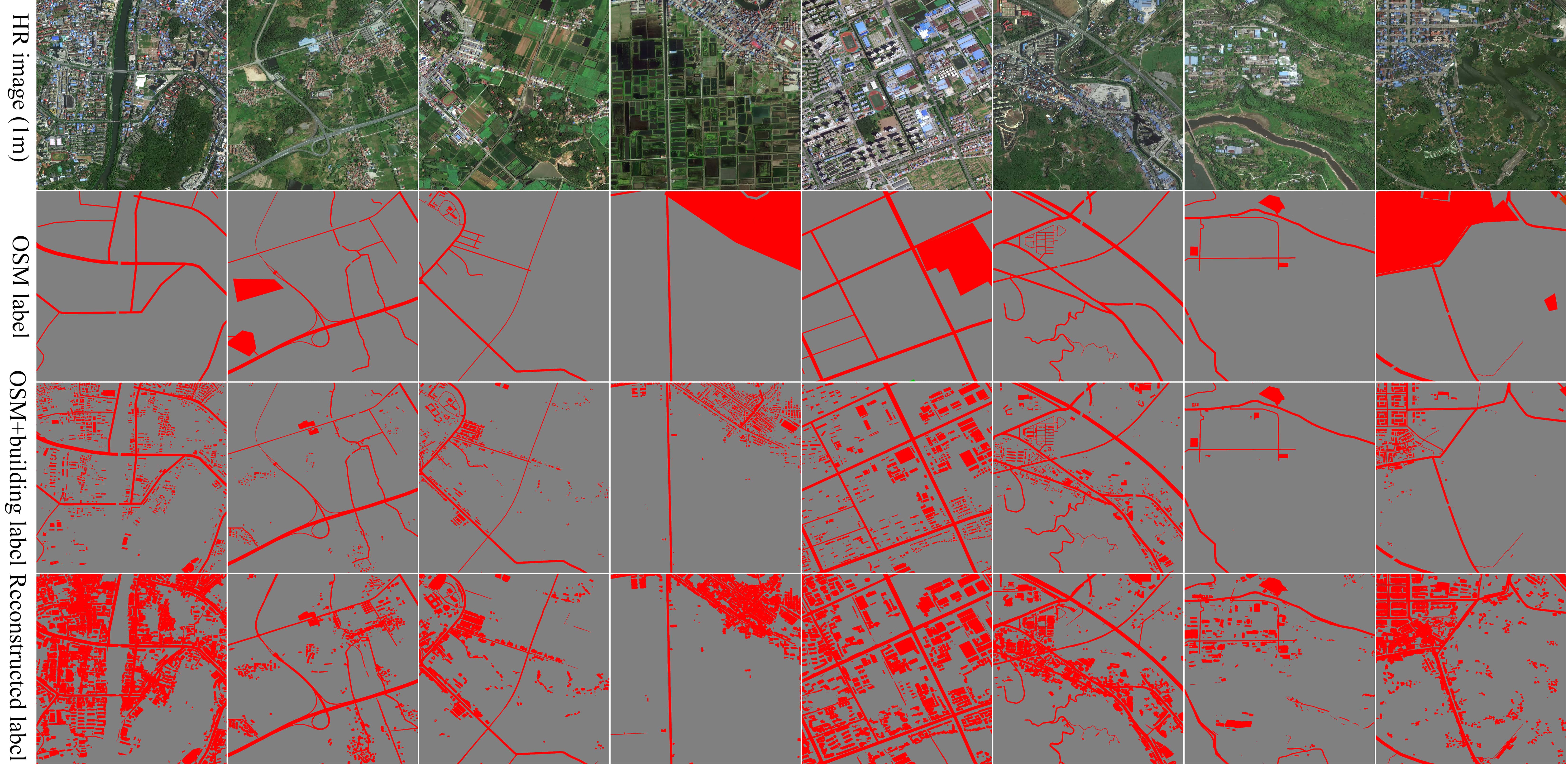}
\caption{\centering The workflow for generating high-resolution impervious surface area (ISA) labels begins with extracting land use/land cover (LULC) categories and road networks from OpenStreetMap. This is followed by incorporating building labels, semi-automated annotation software, and high-resolution RGB imagery to refine the ISA labels further.}
\label{fig:fig3}
\end{figure*}

\noindent\textbf{High-resolution ISA label data:} ISA masks were generated integrating OpenStreetMap (OSM) land use/land cover (LULC) data, road networks, and building footprints \cite{WOS:001363134800001}. The labels in OSM are often incomplete, particularly in many regions of China where significant omissions exist. Moreover, annotations related to ISA are frequently coarse and generalized, lacking spatial precision. Therefore, it is necessary to incorporate high-resolution building footprint data to enhance both the spatial accuracy and applicability of the dataset. Our semi-automated annotation workflow involved generating pseudo labels using the Segment Anything Model (SAM)\cite{Kirillov_2023_ICCV} based on existing OSM and building footprint labels, followed by manual corrections using the Labelme toolkit (v5.6.0). This manual refinement step was necessary because the ISA class encompasses not only roads and buildings, but also features such as airport facilities, plazas, sports fields, and hardened surfaces in industrial areas, all of which require additional manual annotation using the Labelme software. The overall annotation process is illustrated in Fig. \ref{fig:fig3}. The annotated dataset contains approximately 4.26 billion ISA class pixels and 25.33 billion non-ISA class pixels.

\begin{figure*}[!t]
\centering
\includegraphics[width=1\textwidth]{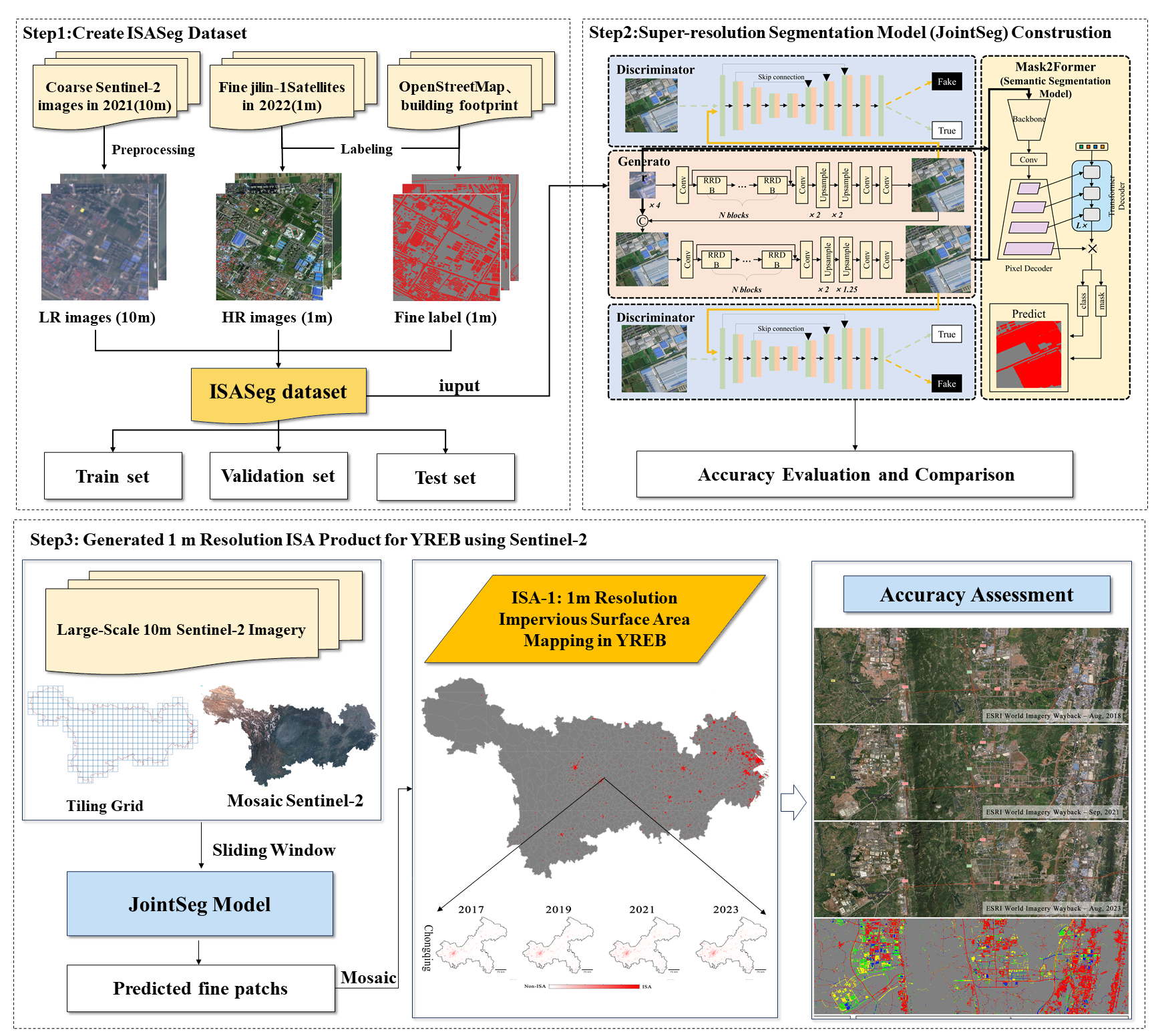}
\caption{\centering Workflow for generating the Artificial impervious surface area (ISA-1, 1m) using Sentinel 2 (10m) satellite images and super-resolution segmentation pipeline.}
\label{fig:fig4}
\end{figure*}

\subsection{Modelling}
The overall JointSeg workflow, as illustrated in Fig. \ref{fig:fig4}, consists of two main stages. First, a dedicated dataset (ISASeg), covering the study area, was carefully constructed to support model training and evaluation. A super-resolution reconstruction model and a semantic segmentation model were trained independently and then integrated into a unified pipeline capable of generating 1m resolution ISA mapping products from input Sentinel-2 imagery at 10m resolution. In the second stage, the validated super-resolution segmentation model was applied to generate multi-year ISA products across the entire YREB, using only Sentinel-2 imagery as input, with quantitative assessments conducted in representative regions to evaluate mapping accuracy.

\begin{figure*}[!t]
\centering
\includegraphics[width=1\textwidth]{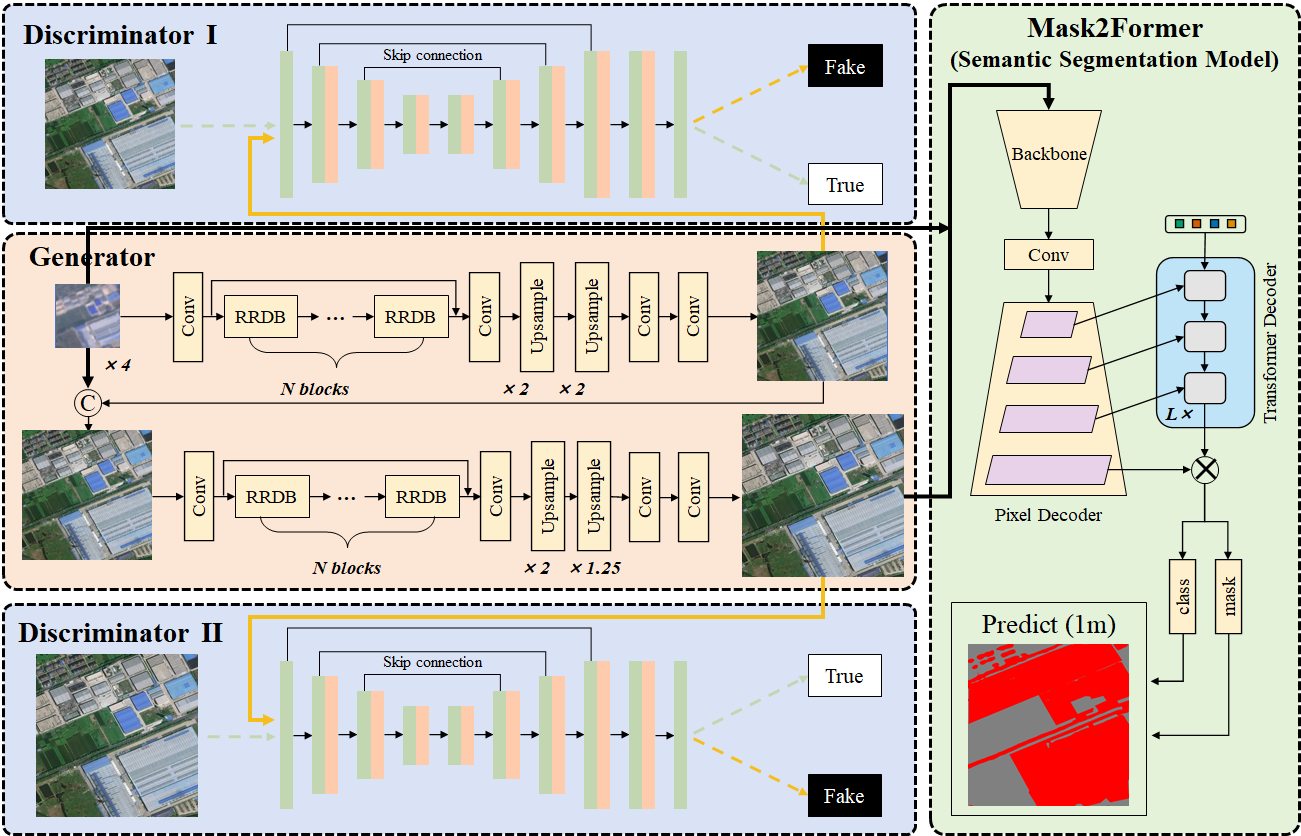}
\caption{\centering The architecture of the super-resolution segmentation (JointSeg) model.}
\label{fig:fig5}
\end{figure*}

\subsubsection{Super-Resolution Model}
Real-ESRGAN\cite{WOS:000739651101110} (Enhanced Super-Resolution Generative Adversarial Network) is a widely recognized deep learning model for image super-resolution. Built on the ESRGAN framework, it features an improved network architecture and enhanced loss functions. The generator adopts a more stable design, effectively capturing fine details while mitigating training instability. Unlike traditional methods that assume ideal low-resolution inputs, Real-ESRGAN is specifically designed to handle real-world degradations,  such as blur, noise, and compression artifacts. By modeling complex degradation processes during training, it achieves robust performance across diverse scenarios. The model integrates perceptual loss, adversarial loss, and pixel loss to enhance detail and texture while preserving overall perceptual consistency. Additionally, its optimized architecture strikes a balance between performance and computational efficiency, enabling real-time inference on consumer-grade hardware. 

However, due to the increasing ill-posedness at higher magnification scales, most existing methods—including Real-ESRGAN—are limited to low upscaling factors (e.g., ×2 or ×4). To address this limitation, this study proposes a progressive super-resolution framework based on ESRGAN, termed Progressive ESRGAN (Prog-ESRGAN), which achieves robust ×10 high-scale super-resolution. As illustrated in the left part of Fig. \ref{fig:fig5}, Prog-ESRGAN takes only the 10 m resolution Sentinel-2 imagery as input. In the first stage, a standard ESRGAN network is employed to perform ×4 super-resolution, where the supervision signal is generated from 1 m high-resolution (HR) imagery degraded through a second-order degradation model. Subsequently, the original Sentinel-2 image is upsampled by a factor of four and fused with the first-stage output, serving as the input to the second-stage ESRGAN network. This two-stage progressive approach ultimately yields a robust ×10 super-resolved result.

Since Prog-ESRGAN is designed to handle a more complex degradation space, a U-Net discriminator with spectral normalization is employed to enhance discriminative capability, allowing it to better manage the intricate outputs generated during training. Moreover, Prog-ESRGAN incorporates two discriminators, each supervising the ×4 and ×10 super-resolution stages respectively, thereby ensuring robust generation throughout the high-scale upsampling process. The training is conducted in two stages. First, we train a PSNR-oriented model using an L1 loss function. The resulting model is referred to as Prog-ESRNet. Subsequently, this pretrained PSNR-oriented model is used to initialize the generator, which is then trained with a combination of L1 loss, perceptual loss, and adversarial loss (GAN loss), yielding the final Prog-ESRGAN model.

\subsubsection{Semantic Segmentation Model} Mask2Former\cite{WOS:000867754201052} unifies semantic, instance, and panoptic segmentation within a single framework. Unlike pixel-wise classification methods, it employs a Transformer-based approach to generate segmentation masks directly, improving both efficiency and adaptability. By leveraging global attention mechanisms, Mask2Former captures contextual dependencies, enhancing its ability to understand complex scenes. The model utilizes query-feature interaction to generate mask representations, with each query encoding semantic or object-specific information. The attention mechanism facilitates interaction between queries and global features, enabling the model to predict masks and corresponding labels in an end-to-end manner. Mask2Former has achieved state-of-the-art performance on benchmarks such as COCO and ADE20K.

\subsubsection{Training Strategy} We first trained the super-resolution model using HR (1m, 1600$\times$1600) and LR (10m, 160$\times$160) image pairs. The generator contains 283.7M parameters, while the discriminator has 70.0M, with a total dataset size of 1732.9GB, providing sufficient data for training such a large-scale model. Using the trained generator, we upscaled 10m LR images to 1m SR and created SR-label pairs for training the Mask2Former segmentation model. Although the super-resolution and segmentation models were trained separately, they are deployed sequentially during inference as a pipeline, producing 1 m-resolution ISA segmentation from Sentinel-2 imagery. Training and inference were conducted in parallel across 10 Nvidia RTX 4090 GPUs.

\section{Results}
\subsection{Super-resolution results}

\begin{figure*}[!t]
\centering
\includegraphics[width=0.8\textwidth]{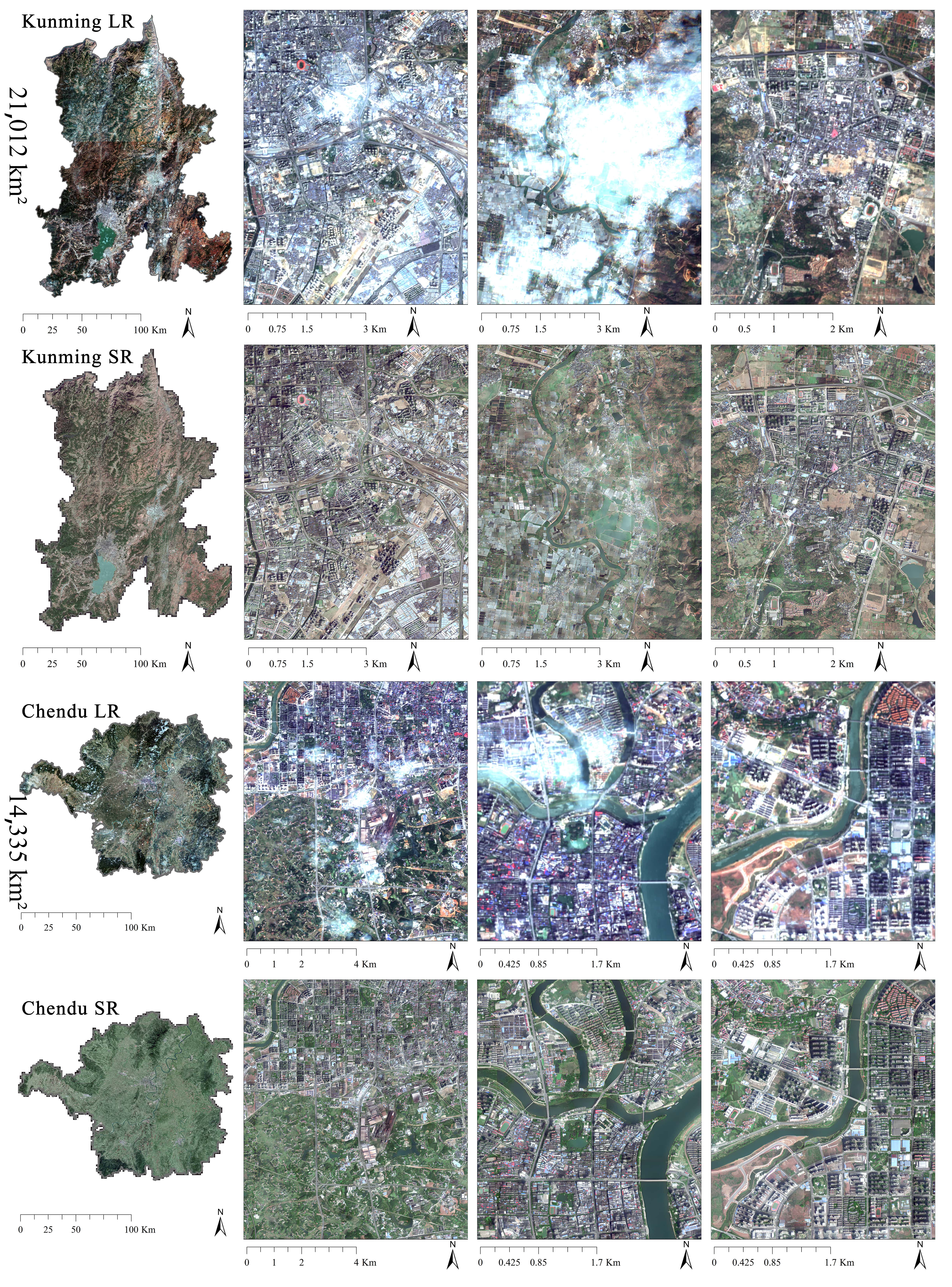}
\caption{\centering Example of LR imagery and SR results for YREB city, demonstrating the model's superior cloud and fog removal capability and generalization.}
\label{fig:fig6_}
\end{figure*}

The core of the JointSeg method lies in its super-resolution and semantic segmentation components, with the super-resolution step being particularly critical. This step is not only responsible for reconstructing high-resolution details from LR imagery but also requires robust capabilities in cross-domain and low-quality image restoration. In practical large-scale mapping applications, factors such as cloud cover, acquisition on different dates, and varying weather conditions introduce significant discrepancies across regions and time periods. These variations severely limit the generalizability of segmentation models. Therefore, the super-resolution module must not only enhance the spatial details of LR images but also normalize stylistic inconsistencies caused by temporal and regional differences in low-quality inputs. This harmonization is essential to ensure better generalization in the subsequent semantic segmentation stage. Fig. \ref{fig:fig6_} presents the super-resolution results for example cities in the YREB region. The training samples covered Kunming, but not Chengdu. The super-resolution results for Chengdu demonstrate the model's generalization capability. The results show that the low-resolution, cloud- and fog-contaminated LR imagery (10m) has been transformed into high-resolution imagery (1m) with exceptional reconstruction quality through the super-resolution model, exhibiting excellent cloud and fog removal effects. Furthermore, when the LR reflectance of a city is heterogeneous, the generated SR is spatially uniform across the entire city, which benefits subsequent segmentation tasks. Notably, the model also performs excellently on Chengdu, a city not included in the training set, highlighting its strong generalization ability.

\subsection{Semantic Segmentation Result}

\begin{figure*}[!t]
\centering
\includegraphics[width=0.8\textwidth]{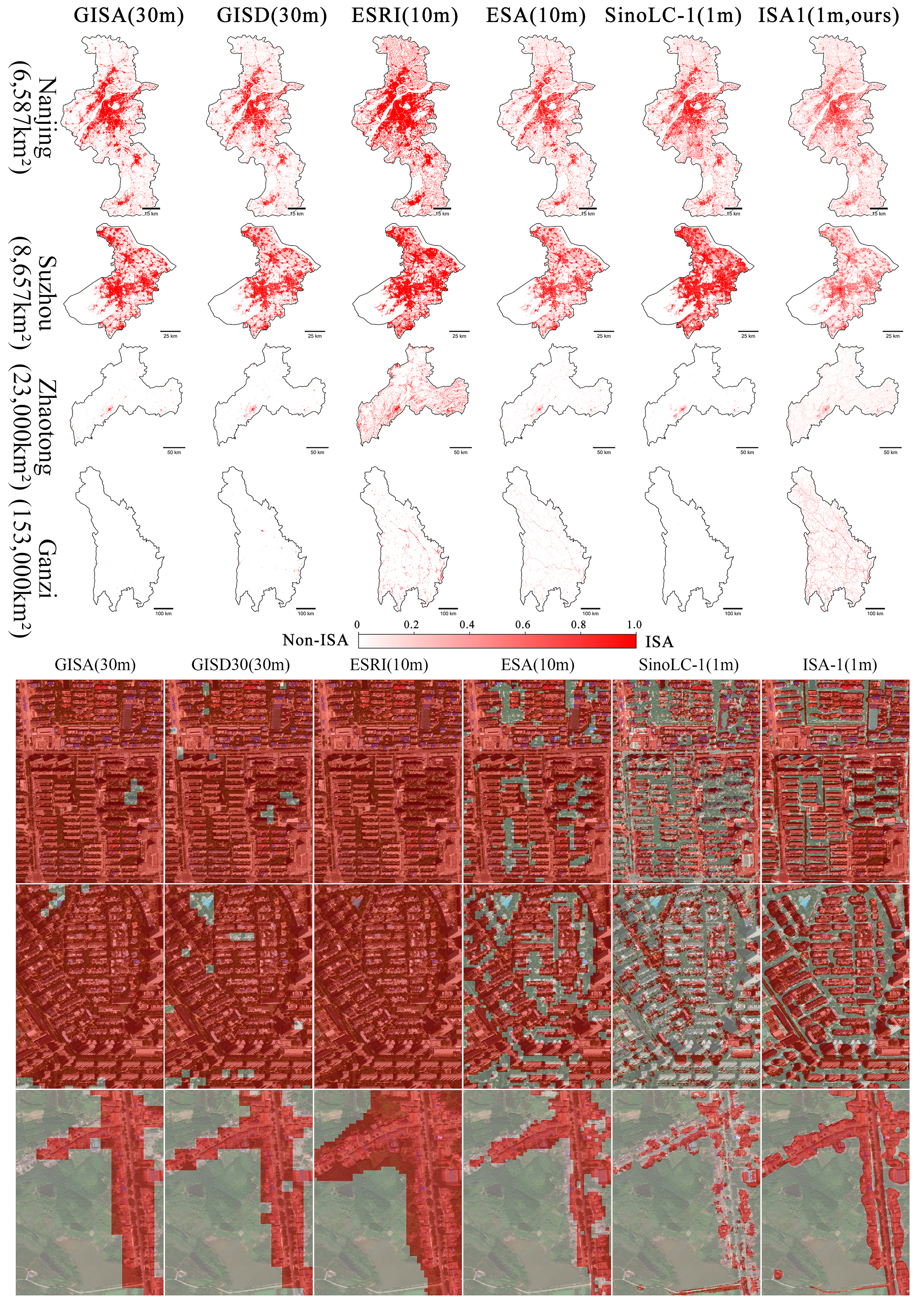}
\caption{\centering Visual comparison of several representative ISA products over typical urban and suburban regions. ISA-1 (ours) exhibits significantly better spatial clarity, particularly in dense built-up areas and fragmented impervious patterns.}
\label{fig:fig7}
\end{figure*}

To comprehensively evaluate the performance of the proposed ISA-1 product (1-meter resolution), we conducted both visual and quantitative comparisons with five representative impervious surface area (ISA) datasets: GISA(30m)\cite{WOS:000679271200001}, GISD30(30m)\cite{zhang2021gisd30}, ESRI (10 m)\cite{karra2021global}, ESA (10 m)\cite{zanaga2022esa}, and SinoLC-1 (1 m)\cite{WOS:001171605700001}. The evaluation includes visual interpretation (Fig. \ref{fig:fig7}) and  a range of evaluation metrics, from detection specificity and sensitivity (precision and recall), to balanced performance (F1-score), spatial agreement (IoU), and global classification success (OA), as summarized in Table \ref{tab:table2}.

\begin{table*}[!t]
\centering
\caption{Comparison of several representative ISA products (\%)}
\label{tab:table2}
\begin{tabular}{ccccccc}
\hline
\multicolumn{1}{l}{}             & Class        & Precision       & Recall          & F1-score        & IOU             & OA                      \\
\hline
\multirow{3}{*}{GISA(30m)}       & Non-ISA      & 0.8899          & 0.8853          & 0.8699          & 0.7942          & \multirow{3}{*}{0.8292} \\
                                 & ISA          & 0.3224          & 0.3326          & 0.2943          & 0.1988          &                         \\
                                 & Mean         & 0.6061          & 0.6089          & 0.5821          & 0.4965          &                         \\
\multirow{3}{*}{GISD(30m)}       & Non-ISA      & 0.8872          & 0.8849          & 0.873           & 0.7951          & \multirow{3}{*}{0.8278} \\
                                 & ISA          & 0.3129          & 0.3463          & 0.3063          & 0.2063          &                         \\
                                 & Mean         & 0.6             & 0.6156          & 0.5896          & 0.5007          &                         \\
\multirow{3}{*}{ESRI(10m)}       & Non-ISA      & 0.9169          & 0.722           & 0.764           & 0.6623          & \multirow{3}{*}{0.7247} \\
                                 & ISA          & 0.2146          & 0.5702          & 0.3003          & 0.198           &                         \\
                                 & Mean         & 0.5657          & 0.6461          & 0.5321          & 0.4301          &                         \\
\multirow{3}{*}{ESA(10m)}        & Non-ISA      & 0.9193          & 0.8671          & 0.8872          & 0.812           & \multirow{3}{*}{0.8491} \\
                                 & ISA          & \textbf{0.5102} & \textbf{0.5814} & \textbf{0.5243} & \textbf{0.3666} &                         \\
                                 & Mean         & 0.7148          & 0.7242          & 0.7058          & 0.5893          &                         \\
\multirow{3}{*}{SinoLC-1(1m)}    & Non-ISA      & 0.8835          & 0.8998          & 0.8787          & 0.8004          & \multirow{3}{*}{0.8266} \\
                                 & ISA          & 0.293           & 0.3566          & 0.2927          & 0.2017          &                         \\
                                 & Mean         & 0.5882          & 0.6282          & 0.5857          & 0.5011          &                         \\
\multirow{3}{*}{ISA-1(1m, ours)} & Non-ISA      & 0.9244          & 0.9997          & 0.9589          & 0.9241          & \multirow{3}{*}{0.9373} \\
                                 & \textbf{ISA} & \textbf{0.9973} & \textbf{0.6176} & \textbf{0.7553} & \textbf{0.6166} &                         \\
                                 & Mean         & 0.9608          & 0.8087          & 0.8571          & 0.7704          &             \\           
\hline
\end{tabular}
\end{table*}

As shown in Fig. \ref{fig:fig7}, in economically developed and highly urbanized cities such as Nanjing and Suzhou, the ISA-1 product demonstrates significantly finer spatial detail compared to other datasets, with notably lower ISA area estimations, particularly in Suzhou and Nantong. This product exhibits superior boundary discrimination along urban peripheries and within internal green spaces and water bodies, indicating enhanced spatial expression capability. It also shows stronger exclusion capacity in green areas and near urban water systems, effectively reducing the false identification of non-ISA regions as ISA. This refinement is not due to decreased detection capability but rather stems from ISA-1's finer resolution and improved classification accuracy for heterogeneous urban landscape features such as parks, rivers, and open spaces, resulting in more nuanced representation at large spatial scales.

In contrast, in mountainous regions such as Zhaotong and Ganzi, which are characterized by fragmented terrain and lower urbanization levels, ISA-1 detects a notably higher number of ISA patches compared to other products. As illustrated in the figure, this product successfully captures low-density settlements, rural roads, and peripheral building clusters distributed across complex topography, which are largely omitted or significantly underestimated by ESA and other lower-resolution datasets. This highlights the limitations of traditional medium- and low-resolution products in detecting ISA in areas with complex terrain and non-conventional urban morphology, and underscores the superior capacity of our high-resolution ISA-1 product to identify small, irregularly shaped ISA patches.

It is also worth noting that among the publicly available products, ESA demonstrates relatively higher accuracy and provides more refined ISA delineation in developed urban areas such as Nanjing, Suzhou, and Nantong. However, despite its 1-meter resolution, the SinoLC-1 product tends to overestimate ISA in large developed cities like Suzhou, showing a pattern similar to the ESRI product, and markedly underestimates ISA in mountainous cities like Zhaotong and Ganzi, resembling the performance of 30-meter products.

As illustrated in the lower section of Fig. \ref{fig:fig7}, the ISA-1 (ours) product significantly improves the delineation of urban structures, particularly in densely built-up areas (top two rows) and suburban regions with fragmented ISA patterns (third row). Compared with coarse-resolution datasets like GISA and GISD30, which suffer from severe blocky artifacts and spatial ambiguity, ISA-1 captures finer spatial features and exhibits a better alignment with the underlying high-resolution imagery. Although the 10-meter products (ESRI and ESA) provide some improvements in capturing ISA details, they still fail to accurately preserve the boundaries of narrow roads and small impervious patches. Despite having the same spatial resolution, SinoLC-1 performs less reliably, showing increased noise and misclassification in both urban and rural environments. In contrast, ISA-1 achieves fine-grained classification of ISA with enhanced spatial consistency.

Table \ref{tab:table2} summarizes the quantitative performance across all datasets. ISA-1 consistently outperforms other products in both overall and category-specific metrics. For the ISA class, ISA-1 achieves the highest precision (99.73\%), recall (61.76\%), and F1 score (75.53\%), with an IoU of 61.66\%, substantially surpassing all competitors. Notably, the average F1 score of ISA-1 reaches 0.7553, and the overall accuracy reaches 93.73\%, indicating its strong capability in accurately detecting both ISA and non-ISA areas.

In contrast, traditional 30-meter datasets (GISA and GISD30) perform poorly on the ISA class, with F1 scores of only 29.43\% and 30.63\%, respectively. This is primarily due to their limited spatial resolution, which hampers the detection of narrow roads and small impervious segments. Among the 10-meter products, ESA (F1 score: 52.43\%) outperforms ESRI (F1 score: 30.03\%), yet still struggles with boundary preservation and class confusion. Although SinoLC-1 benefits from a 1-meter resolution, its higher intra-class variability and false positive rate result in lower precision (29.30\%) and F1 score (29.27\%).

Overall, the ISA-1 product, generated by our proposed super-resolution segmentation approach, demonstrates outstanding performance in both spatial representation and classification accuracy. Its high spatial fidelity and well-trained classification pipeline enable precise, fine-scale ISA mapping, making it highly suitable for applications in urban planning, climate modeling, and sustainable development.

\begin{figure*}[!t]
\centering
\includegraphics[width=0.75\textwidth]{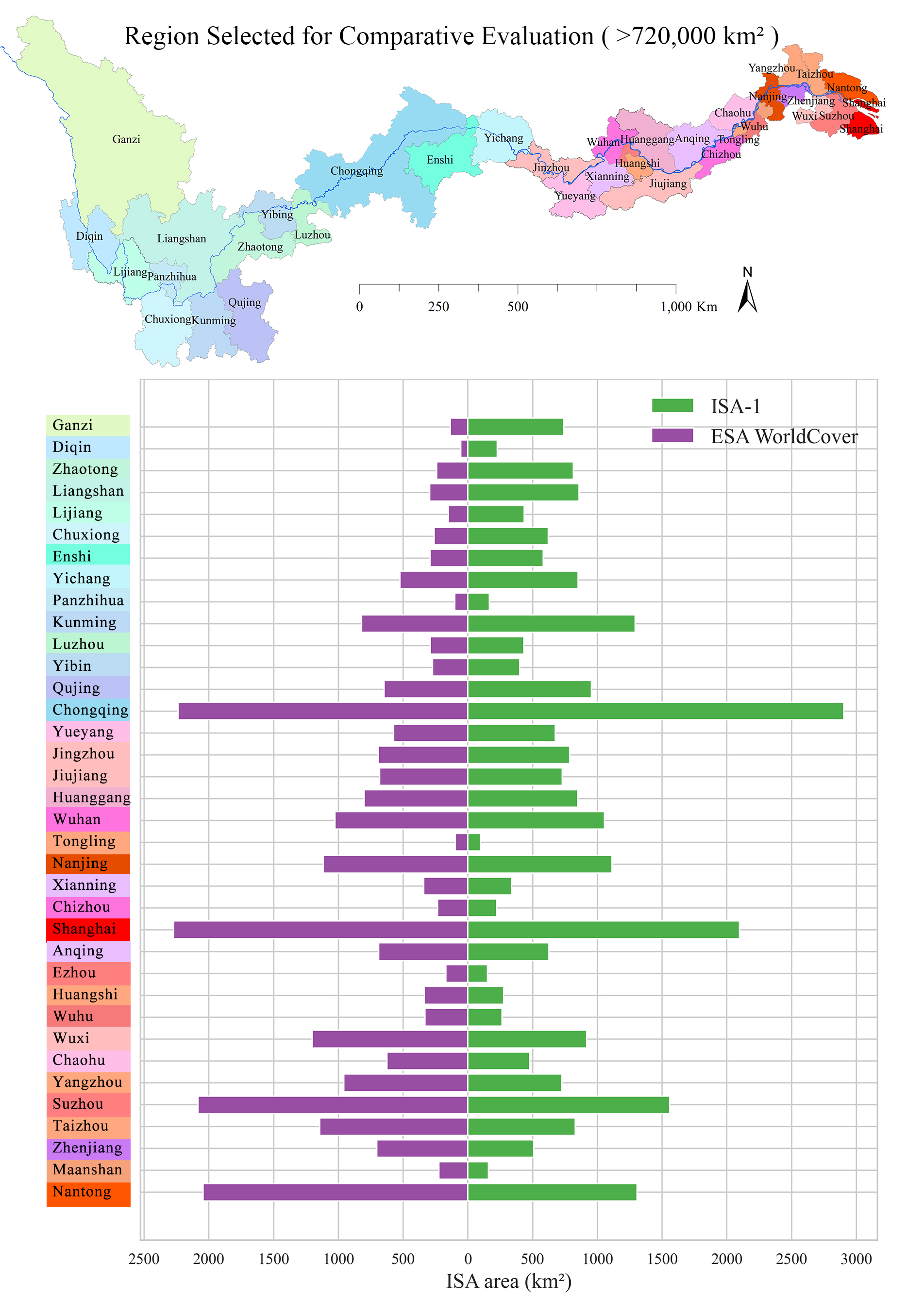}
\caption{\centering Comparison of ISA areas between our ISA-1 product and the ESA-10m product across different regions. In developed cities like Shanghai, Suzhou, Zhenjiang, and Nantong, the ISA area estimated by ISA-1 is slightly lower than that of ESA-10m, while in mountainous-dominated regions such as Chongqing, Diqin, Ganzi, Chuxiong, and Enshi, the ISA area derived from ISA-1 is significantly higher than ESA-10m.}
\label{fig:fig9}
\end{figure*}

\subsection{Large-Scale Regional Comparison}

Given the relatively higher accuracy of the 10m-resolution ESA WorldCover product, a quantitative comparison was conducted to assess discrepancies in ISA estimates between the two datasets. For the 36 cities in the study area, we calculated the ISA class area in the projected coordinate system for both the ESA product and our self-developed 1m resolution ISA mapping product (Fig. \ref{fig:fig9}). 

The results show that in more developed areas such as Shanghai, Wuxi, Yangzhou, Suzhou, and Nantong, our ISA-1 product reports lower ISA areas than the ESA product, with a difference ranging from 0.04\% to 36.15\%. Consistent trends have also been reported by Li et al.\cite{li2022impacts}, who found that the estimated extent of urban areas tends to decrease as spatial resolution improves from 30m to 1m in the context of U.S. cities. This discrepancy primarily stems from the fact that our product can more accurately distinguish between green spaces, water bodies, and other non-ISA land covers within urban areas, thus avoiding the overestimation seen in coarse-resolution products. This high-precision differentiation gives our product a distinct advantage in depicting complex urban land cover structures, especially given the growing demand for detailed urban management and planning. 

In mountainous regions with complex terrain, such as Ganzi, Diqin, Chuxiong, Enshi, Liangshan, and Chongqing, our ISA-1 product identifies 5\%-81.74\% more ISA than the ESA-10m product, indicating significant underestimation by the lower-resolution dataset. This discrepancy highlights the enhanced capability of high-resolution imagery in detecting fragmented ISA features (e.g., rural roads, low-density buildings) in underdeveloped areas. These results demonstrate that improved spatial resolution critically enhances the characterization of dispersed anthropogenic landscapes, particularly in topographically challenging environments where traditional coarser-resolution models have limited sensitivity to small-scale ISA patterns.

\subsection{Analysis of Biennial ISA Products from 2017 to 2023}
\begin{figure*}[!t]
\centering
\includegraphics[width=0.75\textwidth]{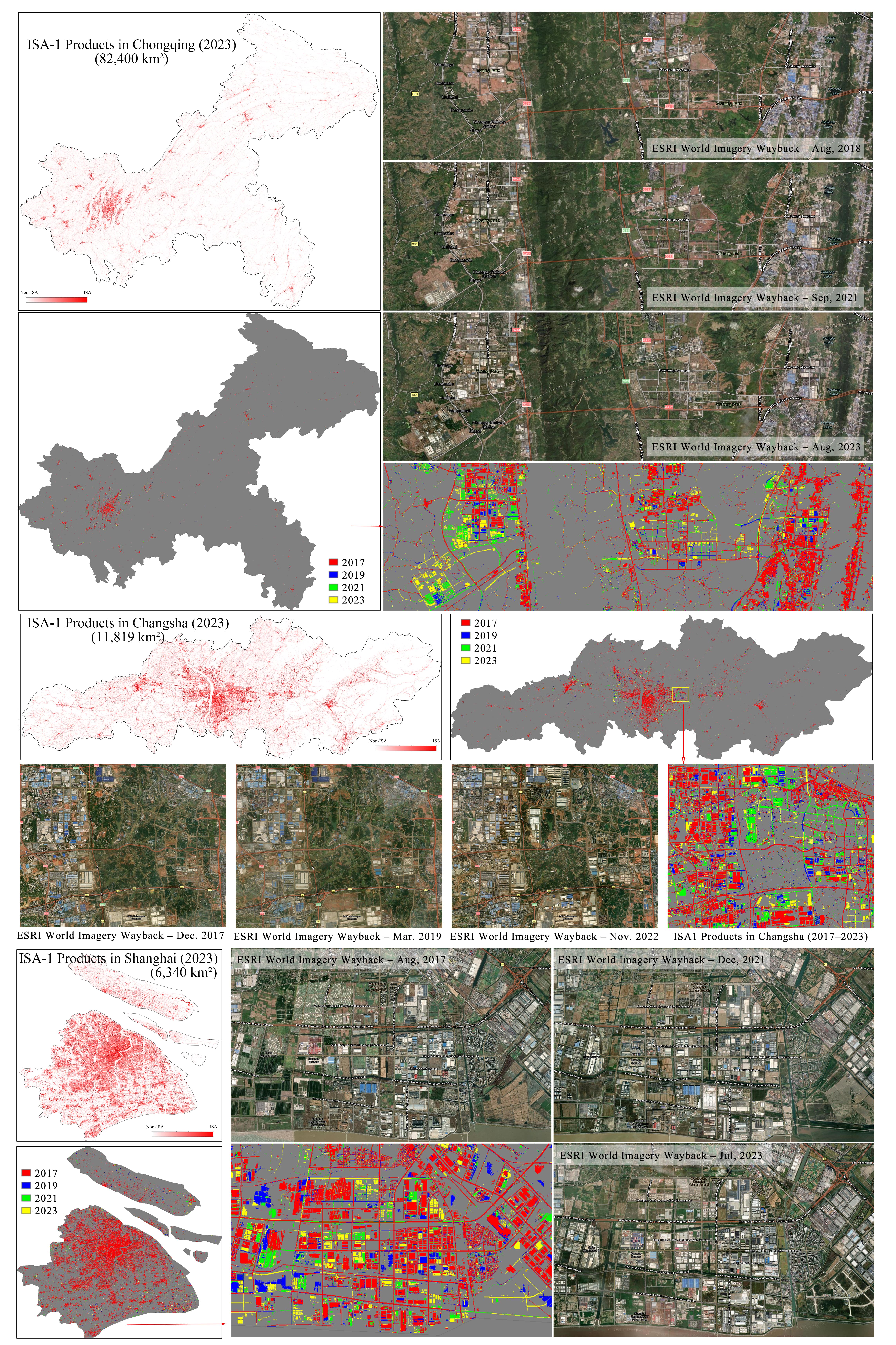}
\caption{\centering Presentation of Biennial Changes in ISA Mapping for Representative Cities in the Upper, Middle, and Lower Reaches of the YREB.}
\label{fig:fig12}
\end{figure*}

Our ISA-1 product reconstructed 1-meter resolution ISA maps for the Yangtze River Economic Belt at biennial intervals (2017, 2019, 2021, and 2023). Fig. \ref{fig:fig12} present the spatiotemporal changes in ISA for representative cities. Table \ref{tab:table3} provides the corresponding biennial ISA areas and change rates for these cities.

\begin{table*}[!t]
\centering
\caption{Inter-biennial Statistics and Change Rates of ISA (km$^2$) in the Study Regions from 2017 to 2023}
\label{tab:table3}
\scalebox{0.55}{
\begin{tabular}{ccccccccccccc}
\hline
Year & \multicolumn{2}{c}{Chongqing} & \multicolumn{2}{c}{Chendu} & \multicolumn{2}{c}{Changsha} & \multicolumn{2}{c}{Wuhan} & \multicolumn{2}{c}{Shanghai} & \multicolumn{2}{c}{Nanjin} \\
\hline
     & Area(km$^2$)    & Change rate    & Area(km$^2$)   & Change rate  & Area(km$^2$)    & Change rate   & Area(km$^2$)  & Change rate  & Area(km$^2$)    & Change rate   & Area(km$^2$)   & Change rate  \\
\hline
2017 & 1836.15      & -              & 1101.14     & -            & 714.17       & -             & 845.29     & -            & 1654.37      & -             & 804.95      & -            \\
2019 & 2310.59      & 25.84\%        & 1407.78     & 27.85\%      & 872.15       & 22.12\%       & 1003.4     & 18.70\%      & 1910.66      & 15.49\%       & 958.16      & 19.03\%      \\
2021 & 2902.48      & 25.62\%        & 1701.35     & 20.85\%      & 1054.4       & 20.90\%       & 1169.79    & 16.58\%      & 2098.19      & 9.81\%        & 1111.97     & 16.05\%      \\
2023 & 3277.37      & 12.92\%        & 1990.95     & 17.02\%      & 1155.87      & 9.62\%        & 1288.13    & 10.12\%      & 2235.48      & 6.54\%        & 1191.02     & 7.11\% \\     
\hline
\end{tabular}}
\end{table*}

As shown in Fig. \ref{fig:fig12}, gradual and small-scale changes in ISA can be observed between 2017 and 2023. By comparing these changes with contemporaneous very-high-resolution imagery (0.6 m), we found that the ISA-1 product faithfully captures fine-grained changes in urban ISA, even when the changes involve only a few isolated buildings. These results further underscore the exceptional performance and consistency of our super-resolution segmentation method, particularly in enabling real-time, high-resolution ISA mapping. The reliability of our method is further validated by its ability to provide accurate and timely ISA assessments, even in complex and dynamic environments.

According to Table \ref{tab:table3}, all six cities exhibited an overall increasing trend in ISA area, though with significant differences in growth rate. Among them, Chongqing in the upper reaches showed the most prominent expansion, from 1,836.15 km$^2$ in 2017 to 3,277.37 km$^2$ in 2023, representing a total increase of nearly 79\%. Notably, the city maintained rapid growth in both the 2017-2019 and 2019-2021 intervals, with increases exceeding 25\% in each period. Middle-reach cities such as Wuhan also showed steady expansion, reaching 1,288.13 km$^2$ in 2023, with an average biennial growth rate of approximately 10\%-18\%. In contrast, lower-reach cities like Shanghai and Nanjing experienced slower ISA growth, with increases of only 6.54\% and 7.11\%, respectively, during the 2021–2023 period—reflecting a trend toward urban saturation and morphological stability.

Overall, cities along the upper, middle, and lower reaches of the Yangtze River exhibit distinct ISA change dynamics characterized by “different development stages—varying expansion speeds—edge-driven growth.” Cities in the upper reaches remain in a stage of rapid urban expansion, middle-reach cities are transitioning toward more intensive and efficient development, while lower-reach cities are entering a phase of urban renewal and functional optimization. These differentiated trends highlight the spatial heterogeneity and gradient nature of urban development along the Yangtze River Economic Belt. They also underscore the importance of region-specific urban planning and differentiated policy implementation to achieve sustainable land use and ecological conservation goals.

\section{Discussion}

\subsection{Performance of Different Super-resolution Models}
To evaluate the performance of our proposed Prog-ESRGAN super-resolution method, we conducted comparisons with several representative approaches, including CNN-based methods such as EDSR\cite{lim2017enhanced} and Real-ESRGAN\cite{WOS:000739651101110}, the Transformer-based method SwinIR\cite{liang2021swinir}, and the stable diffusion-based super-resolution method S3Diff\cite{zhang2024degradation}. For quantitative evaluation, we adopted Peak Signal-to-Noise Ratio (PSNR) and Structural Similarity Index Measure (SSIM) as evaluation metrics. PSNR reflects the pixel-level difference between the reconstructed image and the high-resolution ground truth. A higher PSNR indicates lower reconstruction distortion and better image fidelity. However, PSNR does not align well with human visual perception—it may assign higher scores to blurry images that appear perceptually worse. Moreover, it is insensitive to structural and textural information and highly susceptible to misalignment errors, thus lacking robustness. In contrast, SSIM evaluates image quality from multiple perspectives, including luminance, contrast, and structural similarity. A value closer to 1 indicates greater structural similarity between the compared images and better perceived image quality. SSIM better correlates with human visual perception than PSNR; however, it remains sensitive to image misalignment and may still underperform when assessing images with noticeable artifacts. Therefore, in addition to reporting PSNR and SSIM metrics, we also present visual comparisons of representative super-resolution results to provide a more intuitive assessment of each model’s performance.

\begin{figure*}[!t]
\centering
\includegraphics[width=0.8\textwidth]{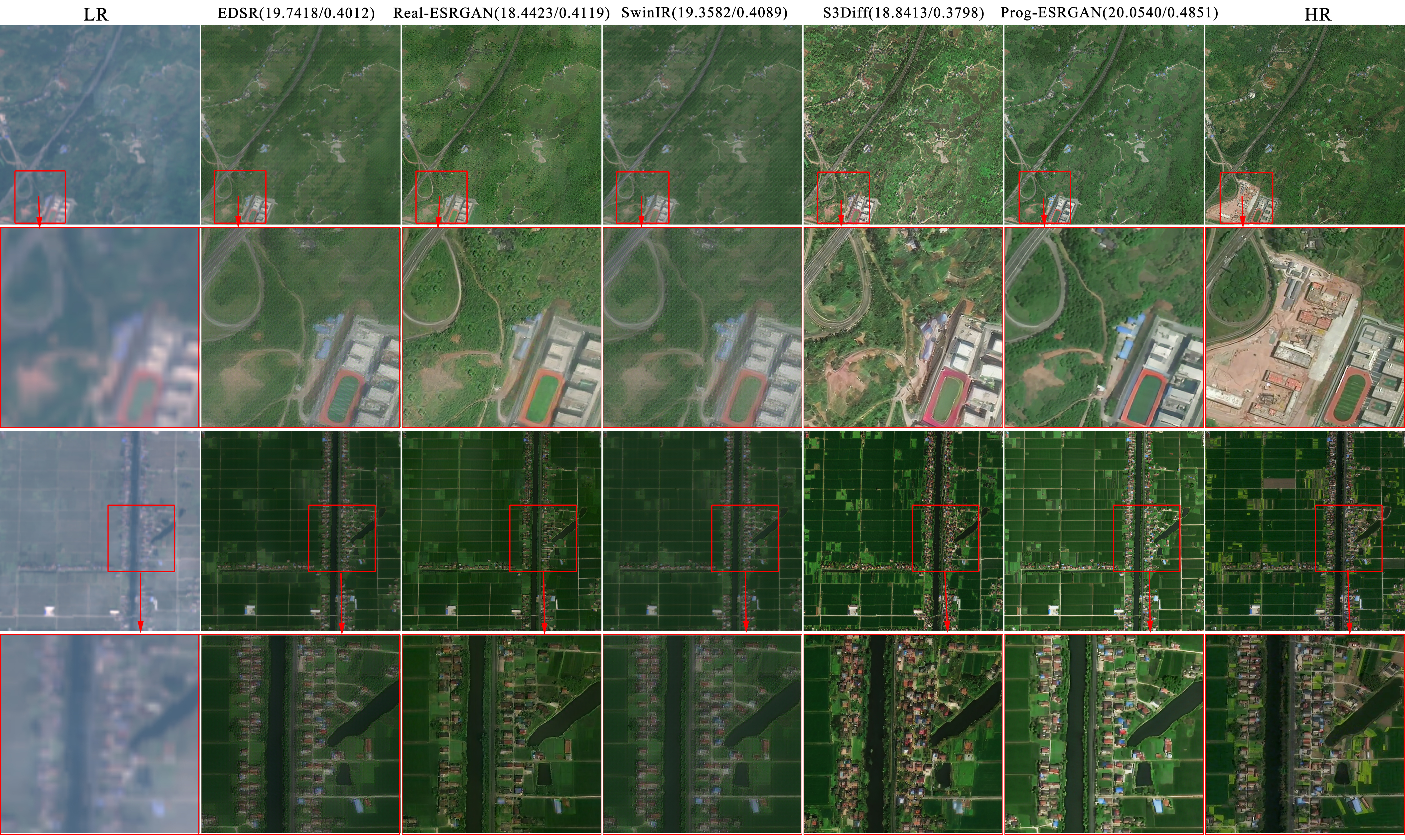}
\caption{\centering Super-resolution results from different models. The values in parentheses indicate the average quantitative performance on the test set in terms of PSNR and SSIM (PSNR/SSIM).}
\label{fig:fig6}
\end{figure*}

Fig. \ref{fig:fig6} presents both qualitative and quantitative comparisons of different super-resolution models across representative remote sensing scenes (values in parentheses denote the average PSNR and SSIM on the test set). Visually, Prog-ESRGAN not only achieves the highest PSNR of 20.8650 and SSIM of 0.4256, but also reconstructs the most visually realistic textures and structural details. For instance, in the first scene, Prog-ESRGAN accurately recovers the outlines of roads and buildings, especially smaller structures that appear noticeably blurred in the LR input and are overly smoothed by EDSR and SwinIR. Although Real-ESRGAN improves texture sharpness, it introduces unnatural artifacts, particularly in vegetated areas. The diffusion-based method S3Diff produces relatively sharp visual outputs, yet fails to preserve the shape integrity of smaller buildings, resulting in severe distortions. Its lower SSIM further reflects poor structural fidelity. Additionally, S3Diff requires approximately ten times the training and inference time compared to Prog-ESRGAN. In contrast, Prog-ESRGAN generates sharper building edges and structural details that closely resemble the HR reference images, outperforming all other methods in both clarity and fidelity. These observations demonstrate that Prog-ESRGAN not only enhances pixel-level accuracy but also substantially improves the reconstruction of high-frequency details, making it a strong candidate for real-world high-resolution remote sensing applications.

Notably, Prog-ESRGAN also exhibits potential for near-real-time land cover mapping based solely on Sentinel-2 imagery. As shown in the second row of Fig. \ref{fig:fig6}, although the HR reference image was acquired in 2022, after significant urban expansion, the super-resolved image generated from 2021 Sentinel-2 data accurately retains the earlier land cover state. This indicates that Prog-ESRGAN can effectively infer fine-scale spatial structures while faithfully preserving the temporal characteristics of low-resolution inputs. Given the global availability and near real-time accessibility of Sentinel-2 imagery, Prog-ESRGAN holds promise for the generation of dynamic, high-resolution land cover maps, without the constraints posed by acquisition delays or limited availability of VHR imagery.

\subsection{Performance of Different Semantic Segmentation Models and Input Modalities}

As shown in Table \ref{tab:table1}, the three representative semantic segmentation models—UNet, Segformer, and Mask2Former—exhibit considerable performance variation in impervious surface area (ISA) recognition under different input modalities (S2, RGB, and RGB+S2), highlighting the critical role of both model architecture and input modality in remote sensing semantic segmentation accuracy.

\begin{table*}[!t]
\centering
\caption{Performance comparison of different semantic segmentation models and input modalities}
\label{tab:table1}

\begin{tabular}{cccccccc}
\hline
Model & Input modality     & Class   & Precision & Recall & F1-score & IOU   & OA                     \\
\hline
\multirow{9}{*}{Unet}        & \multirow{3}{*}{S2(10m)} & Non-ISA & 92.85     & 97.87  & 95.3     & 91.01 & \multirow{3}{*}{91.5}  \\
                             &                          & ISA     & 74.28     & 44.91  & 55.98    & 38.87 &                        \\
                             &                          & Mean    & 83.57     & 71.39  & 75.64    & 64.94 &                        \\
                             & \multirow{3}{*}{RGB(1m)} & Non-ISA & 95.31     & 95.15  & 95.23    & 90.89 & \multirow{3}{*}{91.86} \\
                             &                          & ISA     & 71.93     & 72.63  & 72.28    & 56.59 &                        \\
                             &                          & Mean    & 83.62     & 83.89  & 83.76    & 73.74 &                        \\
                             & \multirow{3}{*}{RGB+S2}  & Non-ISA & 94.42     & 96.97  & 95.68    & 91.72 & \multirow{3}{*}{92.53} \\
                             &                          & ISA     & 79.01     & 66.53  & 72.23    & 56.54 &                        \\
                             &                          & Mean    & 86.72     & 81.75  & 83.96    & 74.13 &                        \\
                             \hline
\multirow{9}{*}{Segformer}   & \multirow{3}{*}{S2(10m)} & Non-ISA & 93.63     & 97.45  & 95.5     & 91.39 & \multirow{3}{*}{91.92} \\
                             &                          & ISA     & 73.41     & 51.56  & 60.58    & 43.45 &                        \\
                             &                          & Mean    & 83.52     & 74.505 & 78.04    & 67.42 &                        \\
                             & \multirow{3}{*}{RGB(1m)} & Non-ISA & 94.8      & 96.43  & 95.61    & 91.58 & \multirow{3}{*}{92.43} \\
                             &                          & ISA     & 76.81     & 69.08  & 72.74    & 57.16 &                        \\
                             &                          & Mean    & 85.81     & 82.76  & 84.18    & 74.37 &                        \\
                             & \multirow{3}{*}{RGB+S2}  & Non-ISA & 94.88     & 96.6   & 95.73    & 91.82 & \multirow{3}{*}{92.65} \\
                             &                          & ISA     & 77.78     & 69.57  & 73.45    & 58.04 &                        \\
                             &                          & Mean    & 86.33     & 83.09  & 84.59    & 74.93 &                        \\
                             \hline
\multirow{9}{*}{Mask2former} & \multirow{3}{*}{S2(10m)} & Non-ISA & 93.58     & 97.34  & 95.42    & 91.25 & \multirow{3}{*}{91.78} \\
                             &                          & ISA     & 72.45     & 51.2   & 60       & 42.85 &                        \\
                             &                          & Mean    & 83.02     & 74.27  & 77.71    & 67.05 &                        \\
                             & \multirow{3}{*}{RGB(1m)} & Non-ISA & 95.71     & 95.78  & 95.75    & 91.84 & \multirow{3}{*}{92.74} \\
                             &                          & ISA     & 75.25     & 74.95  & 75.1     & 60.12 &                        \\
                             &                          & Mean    & 85.48     & 85.365 & 85.425   & 75.98 &                        \\
                             & \multirow{3}{*}{RGB+S2}  & Non-ISA & 92.44     & 99.97  & 95.89    & 92.41 & \multirow{3}{*}{93.73} \\
                             &                          & ISA     & 99.73     & 61.76  & 75.53    & 61.66 &                        \\
                             &                          & Mean    & 96.08     & 80.87  & 85.71    & 77.04 &         \\     
\hline
\end{tabular}
\end{table*}

First, compared to the medium-resolution multispectral Sentinel-2 imagery (S2, 10m), high-resolution RGB images (1m) generated from S2 via our proposed ProgESRGAN model consistently yield better overall performance across all models. This improvement is particularly evident in the ISA class, with noticeably higher recall and F1-score values. For instance, under RGB input, the Segformer model achieves an F1-score of 72.74\% for the ISA class, substantially outperforming the 60.58\% obtained with S2 input. This demonstrates that the spatial detail enhancement achieved by the super-resolution model significantly benefits ISA recognition.

Furthermore, the fused modality (RGB+S2) generally delivers superior comprehensive performance in most models, demonstrating the complementary benefits of integrating multispectral and multi-scale information. In the case of the Mask2Former model, the RGB+S2 input achieves an F1-score of 75.53\% for ISA, outperforming both RGB (75.10\%) and S2 (60.00\%) single-modality inputs. The overall accuracy (OA) also reaches 93.73\%, the highest among all combinations. Overall, Mask2Former yields the best performance with multi-modal input, achieving an average F1-score of 85.71\% and an IoU of 77.04\%, indicating its Transformer-based architecture's strong capacity to leverage both high-resolution spatial and multispectral information. In contrast, although UNet maintains a certain degree of stability under single-modality input, its performance on the ISA class remains consistently lower than the other two models, revealing its limitations in complex object recognition due to weaker feature extraction capabilities. In summary, the combination of multi-modal input (RGB+S2) and advanced Transformer-based architectures (e.g., Mask2Former) significantly enhances ISA segmentation performance. 

\subsection{Advantages of AI-Driven Super-Resolution Segmentation}
Traditional large-scale land cover mapping products (e.g., ESRI Land Cover and ESA WorldCover) are inherently limited by the spatial resolution of their source imagery and typically produce classification outputs at the same resolution as the input data (e.g., 10m)\cite{zanaga2022esa, karra2021global}. The generation of high-resolution (<2m) land cover products has long depended on commercial VHR satellite imagery, such as WorldView and QuickBird. However, the high cost and limited spatial coverage of these datasets have significantly constrained the feasibility of large-scale, dynamic monitoring. In recent years, advances in AI-based image generation and super-resolution techniques—particularly those based on generative adversarial networks and diffusion modal frameworks—offer a promising new avenue for overcoming the resolution and semantic segmentation limitations of medium- to low-resolution remote sensing imagery \cite{WOS:001022958600039, WOS:001269395000002}. These technologies hold promise for establishing a novel paradigm that enables the generation of meter-resolution ISA products from freely available Sentinel-2 imagery at 10-meter resolution.

This study presents the first demonstration of the feasibility of AI-Driven Super-Resolution Segmentation to directly produce 1m-resolution ISA maps from Sentinel-2 data over the YREB, a vast and heterogeneous region of 2.4 million km$^2$ encompassing complex urban agglomerations, mountainous terrain, and dense water networks. The core innovation of our technical framework lies in the joint optimization of super-resolution and semantic segmentation processes: A progressive super-resolution approach based on our customized Pro-ESRGAN model performs a staged upscaling of Sentinel-2 imagery from 10m to 2.5m and ultimately to 1m resolution. This approach effectively addresses the ill-posed nature of large-scale super-resolution in traditional networks and mitigates the excessive smoothing of fine texture details. A multimodal, cross-resolution training strategy is applied using the Mask2Former segmentation model, which combines the original 12-band Sentinel-2 imagery (10m resolution) with the super-resolved 1m RGB imagery. This enables high-fidelity semantic segmentation and enhanced resolution through cross-scale feature fusion. Experimental results demonstrate that our framework achieves significant breakthroughs in ISA mapping, with an F1-score of 85.71\%—a 9.5\% improvement (Table \ref{tab:table1}) over traditional bilinear interpolation-based segmentation methods and a 21.43\%–61.07\% improvement (Table \ref{tab:table2}) over other existing ISA products.

This breakthrough confirms the potential of leveraging the global coverage (5-day revisit cycle) and extensive historical archive (since 2016) of Sentinel-2 to construct a spatiotemporally continuous, meter-scale ISA data cube. Such a capability provides crucial support for high-resolution applications, including urban expansion monitoring and flood risk assessment. The successful implementation of this AI-driven approach has the potential to bring about a significant paradigm shift in large-scale mapping. It paves the way for replacing costly VHR imagery with free medium-resolution data enhanced by AI super-resolution engines, making high-accuracy geographic information services accessible even in developing countries and remote regions. Future work will focus on unsupervised domain adaptation \cite{WOS:001109906200001} and autoregressive pretraining techniques \cite{WOS:001262841000004} to further improve model generalization in complex terrain scenarios.

\subsection{ISA Reduction and Increase Across Regions}

Our comparative analysis reveals pronounced spatial discrepancies between our high-resolution ISA-1 product and the ESA-10m dataset, with these differences manifesting heterogeneously across both developed urban agglomerations and mountainous rural territories. In economically advanced cities such as Shanghai, Wuxi, Yangzhou, Suzhou, and Nantong, the ISA-1 product consistently reports lower ISA estimates—ranging from 0.04\% to 36.15\% lower—than those derived from ESA-10m. This divergence is not anomalous but rather corroborates the spatial resolution effects observed by Li et al.\cite{li2022impacts}, who demonstrated a marked reduction in identified urban extent as spatial resolution increased from 30 m to 1 m in U.S. urban centers. Such findings underscore a critical point: coarser-resolution products tend to misclassify vegetated urban features (e.g., green belts, urban parks) and water bodies as impervious, thereby inflating ISA statistics in dense urban cores. 

Conversely, in rugged and less developed regions such as Ganzi, Diqing, Chuxiong, Enshi, Liangshan, and Chongqing, ISA-1 detects between 5.00\% and 81.74\% more impervious surface than ESA-10m. This significant underestimation by the coarser-resolution product is largely attributable to its inability to resolve fragmented anthropogenic structures—rural roads, dispersed housing, and terraced settlements—that are often embedded within heterogeneous topographic contexts. The improved performance of ISA-1 in these regions highlights the pivotal role of spatial resolution in capturing dispersed and morphologically irregular ISA features. These patterns reinforce earlier arguments on the limitations of medium-resolution data in detecting land surface heterogeneity in topographically complex areas\cite{essd-13-63-2021, WOS:000997229500001}.

Taken together, the ability of ISA-1 to minimize overestimation in urban cores while reducing underestimation in fragmented rural areas positions it as a robust tool for multi-scalar urban and environmental applications. Importantly, the dataset's spatial fidelity enables nuanced assessments of urbanization processes across a socio-ecological gradient—offering insights into both concentrated metropolitan growth and peripheral, dispersed development. This dual sensitivity is especially pertinent in regions like YREB, where urbanization exhibits both high-intensity expansion and fine-grained rural transformations.

\subsection{Limitation and future work}
Although this study constructed a super-resolution segmentation pipeline that enables input 10m resolution imagery to yield 1m resolution segmentation results, the super-resolution and segmentation networks were trained separately. While they can be integrated into a single pipeline for inference, the overall model remains relatively bulky. Related studies \cite{WOS:001257342900001} have shown that, instead of directly using the output of the super-resolution network, utilizing its feature layers can lead to a more effective segmentation process. Moreover, feeding the segmentation results back into the super-resolution module can subsequently enhance the performance of the super-resolution network itself \cite{WOS:000887822800001}. In future research, developing a unified, one-step super-resolution segmentation approach may offer improved computational efficiency. Additionally, due to the loss of texture information for small buildings in low-resolution imagery, super-resolution networks—much like image generation models—may "hallucinate" features, leading to discrepancies in the shape and geographic positioning of the generated buildings compared to their real-world counterparts. Therefore, future studies should consider incorporating fine-grained building and road annotations, along with boundary loss functions, to constrain errors in the super-resolution output using accurate segmentation supervision.

\section{Conclusion}
This study presents a significant advancement in large-scale ISA mapping through the development and application of an AI-driven super-resolution segmentation framework. Leveraging freely available Sentinel-2 imagery (10m resolution), we successfully generated 1-meter resolution ISA products across the YREB, encompassing diverse urban forms and complex topographies over a vast area of 2.4 million km$^2$.

Our approach addresses long-standing challenges in remote sensing-based ISA mapping, particularly the limitations posed by coarse-resolution imagery in accurately delineating heterogeneous ISA. Traditional ISA products such as ESA WorldCover and ESRI Land Cover often overestimate ISA extent in densely urbanized areas due to the inability to distinguish fine-scale urban features from surrounding non-urban elements. In contrast, our ISA-1 product demonstrates superior spatial clarity and segmentation accuracy by integrating a progressive super-resolution pipeline with semantic segmentation via multimodal training. This technical innovation enables robust detection of small and fragmented ISA features, such as rural roads and dispersed settlements, especially in mountainous and underdeveloped regions where conventional methods fail to capture fine spatial heterogeneity.

The ISA-1 product not only improves the granularity and reliability of urban ISA classification but also supports detailed temporal analysis. Our biennial ISA mapping from 2017 to 2023 reveals varying urban expansion dynamics along the upper, middle, and lower reaches of the Yangtze River. For instance, Chongqing experienced a nearly 79\% increase in ISA over the study period, indicative of ongoing rapid urbanization in China’s western interior. In contrast, lower-reach cities like Shanghai and Nanjing exhibited slower growth, reflecting trends toward urban saturation and functional optimization. These spatial and temporal trends underscore the necessity for tailored urban planning strategies and the utility of high-resolution ISA data in supporting sustainable development goals.

Our findings demonstrate that the success of this AI-driven super-resolution mapping paradigm represents a transformative shift in Earth observation workflows. By minimizing dependence on costly commercial VHR imagery and enabling fine-scale mapping through accessible, open-source satellite data, our approach significantly advances the democratization of high-quality geospatial information services, particularly benefiting resource-limited regions.

\subsection*{CRediT authorship contribution statement} 
\textbf{Jie Deng}: data processing, writing- original draft preparation; \textbf{Danfeng Hong}: methodology, supervision, resources, writing-review \& editing; \textbf{Chenyu Li}: technical validation, resources, writing-review \& editing; \textbf{Naoto Yokoya}: supervision, resources, writing-review \& editing.

\subsection*{Declaration of Competing Interest}
The authors declare that they have no known competing financial interests or personal relationships that could have influenced the work reported in this paper.

\subsection*{Data Availability}
This study presents the YREB ISA mapping product (ISA-1), which will be publicly available in the Figshare data repository (\href{https://doi.org/10.6084/m9.figshare.28490204}{https://doi.org/10.6084/m9.figshare.28490204}). It is the first large-scale 1m-resolution ISA mapping product derived from 10m imagery and super-resolution segmentation methods. The data is in GeoTIFF format, with a WGS-84 spatial reference system. The map uses two values: 1 for ISA and 0 for non-ISA. The dataset is compatible with various GIS and remote sensing software, including ArcGIS, QGIS, ENVI, and GDAL, and is freely accessible to all users. 

\section*{Acknowledgments}
This research was supported by the National Key Research and Development Program of China (Grant No. 2022YFB3903401),  the National Natural Science Foundation of China No.32402311, the National Natural Science Foundation of China under Grant 42271350, the China Postdoctoral Science Foundation (Grant No. 2023M743590), the Postdoctoral Fellowship Program of CPSF (Grant No. GZB20230778), the Natural Science Foundation of Beijing Municipality under Grant L222041, and the International Partnership Program of the Chinese Academy of Sciences under Grant No.313GJHZ2023066FN. This research was also supported by the Big Data Computing Center at Southeast University.

\clearpage
\bibliography{reference}

\end{document}